# Variables effecting photomosaic reconstruction and ortho-rectification from aerial survey datasets.

Dr Jonathan Byrne[1], Prof Debra Laefer[1]
[1]Urban Modelling Group, School of Civil Engineering, University College Dublin, Belfield, Ireland.
email: jonathan.byrne@ucd.ie, debra.laefer@ucd.ie

ABSTRACT: Unmanned aerial vehicles now make it possible to obtain high quality aerial imagery at a low cost, but processing those images into a single, useful entity is neither simple nor seamless. Specifically, there are factors that must be addressed when merging multiple images into a single coherent one. While ortho-rectification can be done, it tends to be expensive and time consuming. Image stitching offers a more economical, low-tech approach. However direct application tends to fail for low-elevation imagery due to one or more factors including insufficient keypoints, parallax issues, and homogeneity of the surveyed area. This paper discusses these problems and possible solutions when using techniques such as image stitching and structure from motion for generating ortho-rectified imagery. These are presented in terms of actual Irish projects including the Boland's Mills building in Dublin's city centre, the Kilmoon Cross Farm, and the Richview buildings on the University College Dublin campus. Implications for various Irish industries are explained in terms of both urban and rural projects.

## 1 INTRODUCTION

Ortho-rectified imagery provides a useful tool to civil engineers for planning, boundary mapping, and surveying. It can furnish information sufficiently detailed to derive measured drawings, with the added benefit of containing visual information not typically captured in standard line drawings. However, traditional ortho-rectification is a slow and labour-intensive process. Once aerial imagery has been gathered, a surveyor must collect ground control points (GCP) visible in the overhead images. This involves either using a global positioning system (GPS) with a real time kinematics system or a total station throughout the survey area. While both can provide excellent accuracy, the costs to do so are high.

Alternatively, there are automatic techniques for matching and merging images. The two primary approaches to automatically generating orthomosaics are photo-stitching and Structure From Motion (SFM). Photo-stitching combines multiple images with overlapping fields of a view to create a high resolution image. While the process is very fast, lighting variance and image alignment can be problematic. In contrast, the newer SFM approach operates on a principle similar to stereo vision to generate a three-dimensional (3D) scene model. Although SFM is much more computationally expensive, it generates a 3D model, which can be used to generate an ortho-rectified image; photo-stitching cannot.

UAV photography differs from traditional satellite or aerial imagery in that it is usually captured at a much lower height: normally between 30 m and 100 m. Accordingly, specific factors complicate the ortho-rectification process. Effects such as the parallax caused by terrain features (not normally an issue when images are taken at a distance greater than 500 m), become magnified at such a range. Additionally, the number of large-scale features in each image is reduced, which further complicates finding suitable keypoints with which to align the separate images. Keypoints are unique points in an image that can be matched to a corresponding point in another image.

This work investigates the effect of image capture height on the ortho-rectification process for civil engineering projects by comparing three different datasets and three processing methods: two different photo-stitching approaches (Hugin [1] and Microsoft image composite editor [2]) and SFM in the form of VisualSFM [3] and by quantifying which technique(s) works best for different surveying objectives.

By controlling multiple variables in the aerial datasets (lighting, overlap, height, and rotational variation) the sensitivity of the various approaches can be evaluated with respect to identifying an optimal approach for a given project – particularly with respect to achieving high quality, ortho-rectification results.

## 2 AUTOMATIC ORTHORECTIFICATION TECHNIQUES

### Photo-stitching

Combining images through aligning and stitching them into seamless photo-mosaics is one of the oldest and most widely applied algorithms in computer vision [4].

The basic principle has four steps: keypoint detection, image registration, calibration, and blending. Keypoint detection applies an algorithm such as SIFT [5] or Harris corner detection [6] to generate points of interest for a given image. Keypoints are groups of pixels that have a specific quality, such as being an edge or a corner that allows them to be identified in different images at various angles and scales. They must have a well-defined location and be robust to changes in lighting and viewpoint. Once keypoints have been obtained for all the images, a robust matching algorithm, such as Random Sampling and Consensus (RANSAC) [7] is applied to find inlier matches between images.

Calibration involves two steps. First, the image distortion due to the lens and camera are modelled and geometrically corrected. The images are aligned by applying translation, rotation, and scaling transformations, as described by a homography matrix [8]. A homography matrix has eight parameters or degrees of freedom that can be computed using a direct linear transform and Singular Value decomposition [8]. The second step involves blending the images. Blending combines the remapped images into a single output projection and adjusts the colour between images to compensate for exposure difference, thus minimising the seams between images.

An advantage of photo-stitching is that it is fast, as it normally uses only a small number of keypoints. However, the terrain must be flat and the images well aligned. Any stitching error is perpetuated and accumulated throughout the process. Additionally, as ortho-rectification occurs, the process cannot generate results suitable for deriving accurate measurements. In this work two photo-stitching approaches [Hugin and Microsoft's image composite editor (ICE)] are compared to each other and to Structure from Motion.

### 2.2 Structure from Motion

Structure from motion (SFM) is a methodology for automatically reconstructing 3D models from a series of two-dimensional (2D) images when there is no a priori knowledge of the camera location and/or direction. SFM is a robust and automatic technique that has found widespread adoption in a number of fields [9] and has been used to reconstruct 3D models from sets of images obtained from different sources, such as Snavely's work constructing 3D models of famous architectural sites from images downloaded from the internet [10]. SFM is a fully automatic procedure that uses several computer vision techniques and feature detection algorithms to simultaneously solve the 3D structure of a scene and the viewing parameters to recreate a 3D model.

The SFM process has four steps: (1) feature detection, (2) alignment, (3) bundle adjustment, and (4) reconstruction. Similar to image stitching, the first step is to find keypoints in the image but the number of keypoints generated is usually much higher, approximately 10 times as many keypoints are required. As with image matching, alignment is achieved by using RANSAC for finding matches and removing outliers. Next, bundle adjustment [11] refines a visual reconstruction to produce jointly the optimal 3D structure and the viewing parameters. The "bundle" refers to the bundle of light rays leaving each 3D feature and converging on each camera centre. The bundles are optimally adjusted with respect to both feature and camera positions. Bundle adjustment must occur after outlier removal, as the process is sensitive to noise.

The result of bundle adjustment is a sparsely populated point cloud of the 3D scene. The final reconstruction step thickens this cloud by using the image data and interpolating more points in the scene. This technique is called multi-view stereopsis (MVS) [12] and uses points that have already been matched. MVS interpolates points using the previously identified points from the image data. Once additional points have been generated, visibility filters are applied to remove invalid points, such as points that are occluded by other parts of the point cloud.

SFM is much more computationally expensive technique than photo-stitching but can provide a greater level of accuracy. Perspective information in the images can be used to build 3D models within SFM. Each model is then flattened to produce a 2D image that is free from perspective distortion and suitable for measurement.

## 3 EXPERIMENTAL SETUP

The intention of this work is to examine how the buildings on the ground affect surveys from lower altitudes. Specifically what features are accurately ortho-rectified by the two different methodologies

In order to provide a consistency between datasets, each was acquired at a specific height and has an 80% vertical and 60% horizontal overlap. This was accomplished using the Pix4D capture mapper software that automatically generates the waypoints and images for a given survey area. The images are acquired vertically downward with the gimbal facing towards earth to ensure there is no variance due to turbulence. A serpentine (back and forth) path was flown for the missions, rather than a zig-zag flight path. All flights were conducted on the same day and within minutes of each other so as to control for lighting and cloud occlusion, which are known to generate problems in image processing.

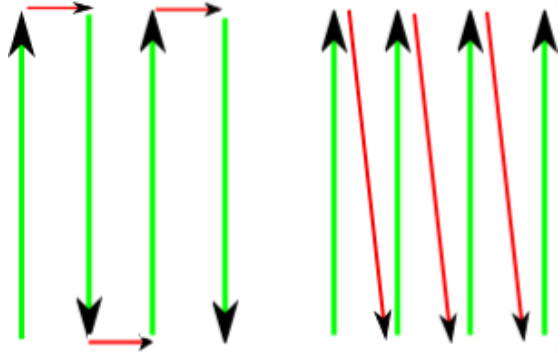

Figure 1 A serpentine flight path versus a zig zag flight path

An unmanned aerial vehicle in the form of the Phantom 3 Professional was used to gather the datasets. It has a self-stabilising 3-axis gimbal that ensures the nadir images are correctly oriented. The Sony Exmor camera used in these experiments provides images with a 94 Field of view (FOV) and a small amount of rectilinear distortion. The resulting images are 4000 by 3000 pixels. At a height of 50 m, this translates into a ground sampling distance of 2 cm/pixel.

### 3.1 Height combinations

Ensuring an 80% overlap at distinctive heights results in a different number of images per dataset. Capturing a 500 m² area at 50 m results in 30 images, whereas capturing the same area at 30 m results in 160 images. Lower heights result in much more data and a significantly better ground sampling distance (GSD); the GSD is the distance between the centre of two consecutive pixels measured on the ground. A better GSD complicates the matching process by introducing a more extensive dataset.

Different height combinations were sampled for the three different projects. The Richview site and the Kilmoon Cross Farm site were captured at 3 elevations from as little as 30 m to as much as 70 m. Due to aviation authority restrictions only a single height was used for the Boland's Mills. The images were captured from a height of 85 m in height, which was 30 m above the tallest building.

### 3.2 Software settings

The camera settings were detected automatically from metadata in the images. The focal length of the Exmor camera was 3.61 mm and the focal length multiplier was 5.54. The built in keypoint detector (CPFind) was used, and the geometric solver optimised the camera position and translation.

Image Composite Editor [2] allows the type of camera motion to be specified. The motion type was set to planar motion image to a serpentine sequence. The auto-overlap settings were used (80% horizontal, 80% vertical, and 20% search radius), as the authors previously found that specifying the exact overlap reduces the searched area and produces worse results.

The default settings were used for VisualSFM [3], and siftGPU [13] was used for keypoint detection. RANSAC was employed for finding the inliers and Parallel Bundle Adjustment (PBA) bundle adjustment. Dense point cloud reconstruction was conducted using the clustered Multi View Stereo (CMVS) library, all of which is standard with VisualSFM.

## 4 DATASETS

### 4.1
Three controlled datasets were used in this study. The areas included the Richview portion of the University College Dublin campus, a glasshouse on the Kilmoon Cross Farm, and the Boland's Mills, a 19th century industrial building in Dublin's city centre. The choice of sites and their respective features are discussed in detail below.

### Richview Buildings

The Richview building complex is home to the University College Dublin's School of Architecture, Planning and Environmental Policy. This portion of the campus has been extensively surveyed by total station and by terrestrial LIDAR, which allows verification of the accuracy of the orthomosaic. The Richview site was chosen because it has buildings with numerous, varied features, such as corners and edges of complicated building profiles and little noise from reflections. The building heights ranged between 6 and 30 m. The images were taken from 30, 40 and 50 m above ground level (AGL). There were few reflective surfaces to introduce noise into the algorithm.

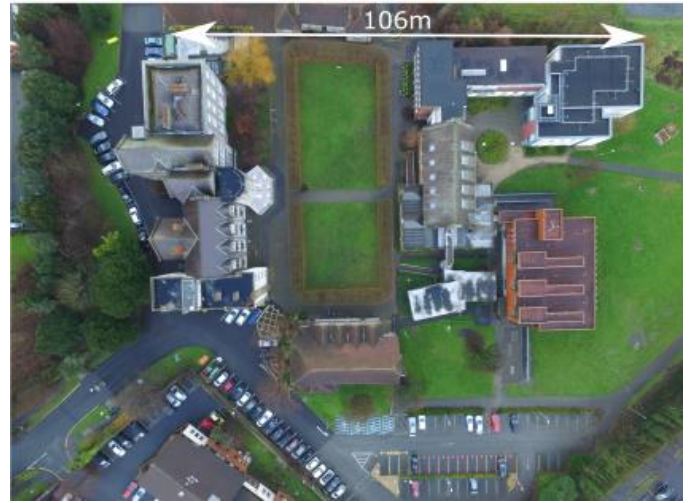

Figure 2 Richview Buildings

### 4.2 Kilmoon Cross Farm

Due to the regulations placed by the Irish Aviation Authority (IAA) on unmanned aerial vehicles (UAVs), permission to fly them outside controlled airspace is easier to obtain. Kilmoon Cross farm is outside the control of Dublin air traffic control. In such cases, only the property owner's permission was necessary. The glasshouses provided a particularly difficult challenge for ortho-rectification, as the structures were uniform and glass is highly reflective. Reflections generate transient keypoints that change from image to image. Uniformity in a structure results in improper matching as different homography matrices produce seemingly good results through incorrect alignment. The data were captured 30 m, 50 m, and 70 m above the ground.

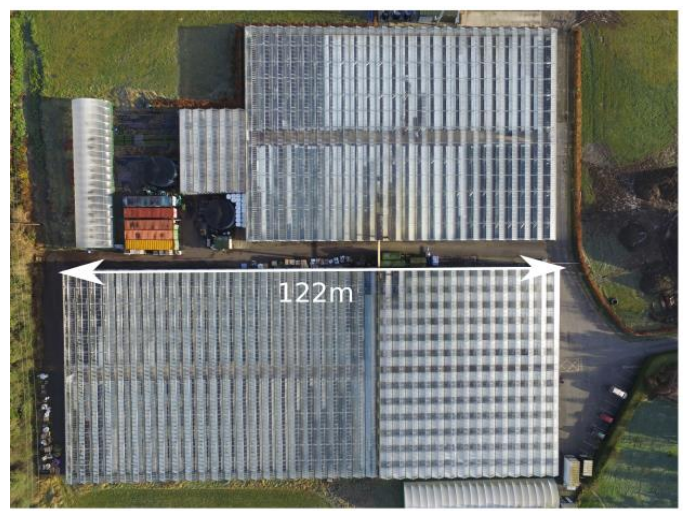

### 4.3 Boland's Mills

Boland's mill's is a site of historic significance in Dublin city centre. The first structure was built in 1830 and it had further concrete silos that were built between the 1940s and the 1960s. The mill ceased operation in 2001 and is now undergoing a 150 million euro conversion into office spaces and residential housing. As several of the buildings in the complex are listed in historic registries, an aerial survey was conducted to create a permanent digital document of the site.

Boland Mills has several tall structures (55 m) that introduce a large amount of parallax between images. The survey was conducted at a height of 85 m, which was 30 m above the rooftop of the highest building. Data could only be captured from a single height due to the congestion of the built environment and the proximity to the Dublin airport. These factors controlled both the minimum and maximum possible flight altitude.

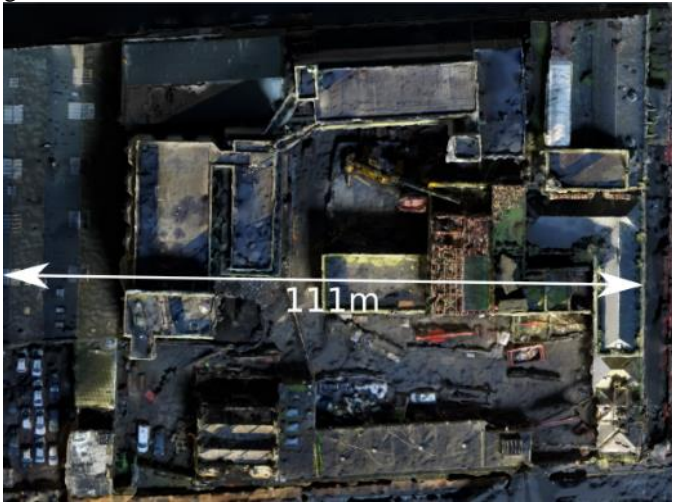

Figure 3 Boland's Mills

## 5 RESULTS

The results for the three experiments are shown below. Due to space limitations only the orthomosaic results for the best result of each is shown. The other images can be viewed online at [**13**]. In order to highlight errors in the reconstruction, different colours are used. Red indicates that the overlap removed information and yellow highlights distortion due to parallax.

### 5.1 Hugin

Hugin failed on all the datasets to produce a coherent orthomosaic. The settings were double checked on the example image sets and with aerial datasets taken at a greater height (280 m-500 m). The approach was also tested with smaller samples of the datasets where it was able to match shorter sequences of images, but it failed completely when being applied to more than 5 images in a non-linear translation. Accordingly, Hugin performed worst of the three methods

### 5.2 ICE

ICE generated coherent orthomosaics for all of the datasets but each contained some level of distortion. The Richview dataset highlighted the difficulty caused by height. At 30 m, all of the buildings, except the rectangular building in the lower right of the image, were distorted. As the height increased, the distortion was reduced, but there was still misalignment on the tallest building in the upper part of the image. One issue was that the images chosen for stitching sometimes contained significant parallax, i.e., there is a difference in position when viewed from two different lines of site. An example of this is the gable wall of a building visible in un-rectified imagery. if the image was truly orthographic, only the roof would be visible and not the side walls of a structure.

As the algorithm is only concerned with merging the key points, it has no ability to correct for the parallax in the image.

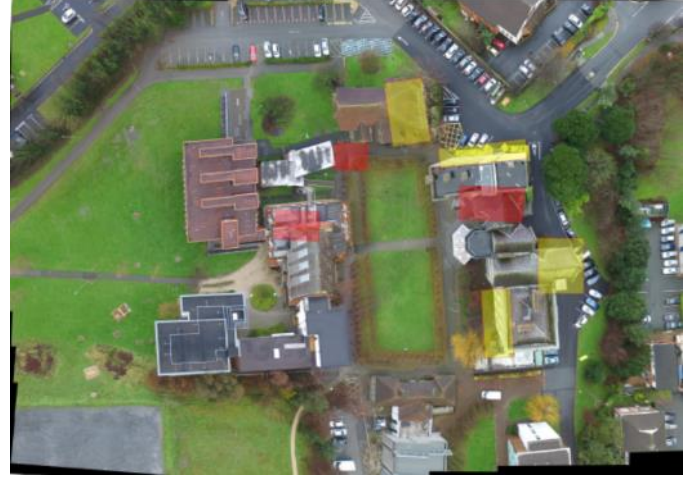

Figure 4 ICE reconstruction for 60m Richview dataset

ICE generated the best result for the Kilmoon dataset. Although there was some distortion, particularly in areas where the structures were very similar, overall the reconstruction was accurate. Error propagation was visible in the glasshouses in the lower half of the image. Once it had aligned the images incorrectly, the error continued for the rest of the glasshouse, as highlighted by the red lines.

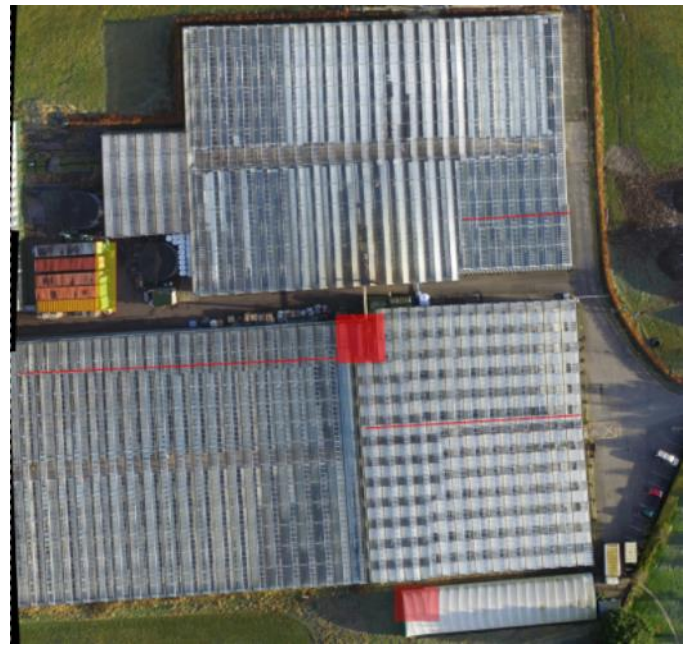

Figure 5 ICE reconstruction for the 70m high Kilmoon dataset

The biggest challenge for ICE was the Boland's mills dataset. There was significant loss of detail on the tallest buildings in the upper part of the scene. In several places, the buildings were either merged together or completely occluded, as highlighted in red. The gable end of several of the building is also shown in yellow as they should not be visible in an ortho-rectified image.

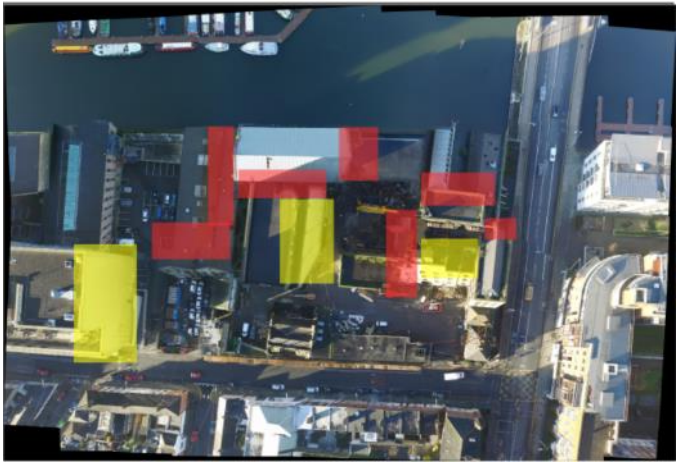

Figure 6 ICE results on 30m Boland's mill dataset.

### VisualSFM

VisualSFM generated coherent mosaics for two of the three datasets with high accuracy and little distortion. The Richview dataset generated a single model and accurate results at all heights with little distortion or parallax effect on the buildings.

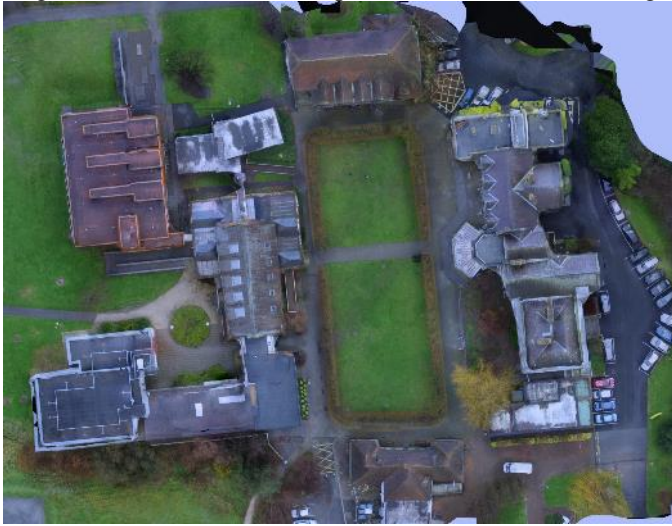

Figure 7 VisualSFM result for 60m Richview dataset

VisualSFM failed completely on the Kilmoon dataset and generated multiple separate models for each height. Because glasshouses constitute a similar pattern, matching keypoints is problematic due to the high number of false positives. There is also the issue of reflection in the glass, which generates false keypoints, thereby further confusing the algorithm. Interestingly, the best result was generated by the 30 m dataset, which may have contained more small scale features for alignment. The resulting model also contained a curvature that could have been caused by the lens distortion in conjunction with the poor quality of the model.

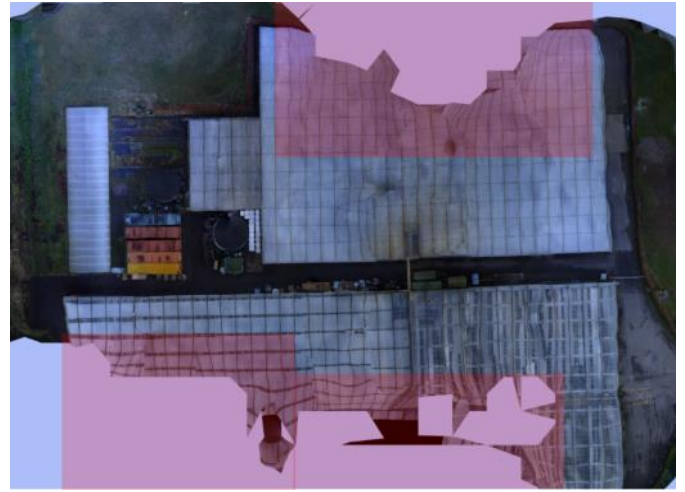

Figure 8 VisualSFM result for 30m Kilmoon dataset

The Boland's mills dataset generated a single consistent model that only contained one area of missing information and one small area showing part of the gable end of the building. The results were further verified by comparing the 3D model with survey results and were found to be accurate to a centimetre level.

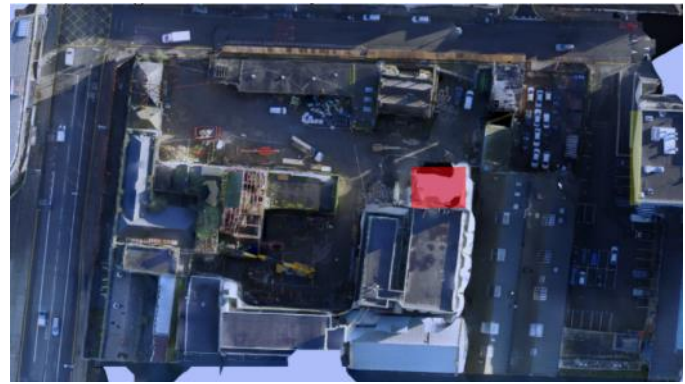

Figure 9 VisualSFM result for 30m Boland's mills dataset

## 6 CONCLUSIONS

Photo stitching works well with low parallax datasets, but it introduces significant distortion and errors when stitching images gathered at a low altitude. While Hugin has previously been shown to work well for aerial datasets gathered between 280 m and 500 mm and for generating panoramas, it failed to generate an orthomosaic for any of the datasets. In contrast, ICE generated orthomosaics for all the datasets and generated the best result for the Kilmoon dataset. These results show that photo stitching is not accurate at low altitudes and only applicable to sites with little parallax, uniformity, or noise from reflection, such as agricultural applications or river mapping.

Conversely, structure from motion generated highly accurate models for built up urban environments. The parallax of the buildings generates height information that is used to create 3D models, and the high level of variation allowed for accurate feature matching. Notably, however, SFM could not create reconstructions of areas where there was little parallax, high uniformity or significant noise. Interestingly it performed better at lower altitude in these cases, which highlights that SFM should be applied for low-level scanning with UAVs.

## ACKNOWLEDGMENTS

This work was funding by IRC grant number GOIPD/2015/125, Science Foundation Ireland 13/TIDA/1274 and Irish Geological Survey 2015-SC-042. We would like to thank ARUP for allowing us to conduct the survey of Boland's Mills and Andrea McMahon for her unceasing support.